\documentclass[sn-aps]{sn-jnl}

\usepackage{graphicx}%
\usepackage{multirow}%
\usepackage{amsmath,amssymb,amsfonts}%
\usepackage{amsthm}%
\usepackage{mathrsfs}%
\usepackage[title]{appendix}%
\usepackage{xcolor}%
\usepackage{textcomp}%
\usepackage{manyfoot}%
\usepackage{booktabs}%
\usepackage{algorithm}%
\usepackage{algorithmicx}%
\usepackage{algpseudocode}%
\usepackage{listings}%
\usepackage{tabularx}
\usepackage{changes}
\usepackage{xcolor}

\begin{document}

\title[Article Title]{Stable and Interpretable Deep Learning for Tabular Data: Introducing InterpreTabNet with the Novel InterpreStability Metric}


\author[1]{\fnm{Shiyun} \sur{Wa}}\email{shiyun.wa23@imperial.ac.uk}  

\author[2]{\fnm{Xinai} \sur{Lu}}\email{xinai.lu@ucdenver.edu}  

\author*[3]{\fnm{Minjuan} \sur{Wang}}\email{minjuan@cau.edu.cn}  

\affil[1]{\orgdiv{Applied Computational Science and Engineering}, \orgname{Imperial College London}, \orgaddress{\street{South Kensington Campus}, \city{London}, \postcode{SW7 2AZ}, \state{London}, \country{UK}}}

\affil[2]{\orgdiv{Economics}, \orgname{University of Colorado Denver}, \orgaddress{\city{Denver}, \postcode{80202}, \state{Colorado}, \country{USA}}}

\affil*[3]{\orgdiv{College of Information and Electrical Engineering}, \orgname{China
Agricultural University}, \orgaddress{\city{Beijing}, \postcode{100083}, \state{Beijing}, \country{China}}}

\abstract{
As Artificial Intelligence (AI) integrates deeper into diverse sectors, the quest for powerful models has intensified. While significant strides have been made in boosting model capabilities and their applicability across domains, a glaring challenge persists: many of these state-of-the-art models remain as ``black boxes’’. This opacity not only complicates the explanation of model decisions to end-users but also obstructs insights into intermediate processes for model designers. To address these challenges, we introduce InterpreTabNet, a model designed to enhance both classification accuracy and interpretability by leveraging the TabNet architecture with an improved attentive module. This design ensures robust gradient propagation and computational stability. Additionally, we present a novel evaluation metric, InterpreStability, which quantifies the stability of a model's interpretability. The proposed model and metric mark a significant stride forward in explainable models' research, setting a standard for transparency and interpretability in AI model design and application across diverse sectors. InterpreTabNet surpasses other leading solutions in tabular data analysis across varied application scenarios, paving the way for further research into creating deep-learning models that are both highly accurate and inherently explainable. The introduction of the InterpreStability metric ensures that the interpretability of future models can be measured and compared in a consistent and rigorous manner. Collectively, these contributions have the potential to promote the design principles and development of next-generation interpretable AI models, widening the adoption of interpretable AI solutions in critical decision-making environments.
}


\maketitle

\section{Introduction} \label{sec1}
AI models, with their profound capacity to process and analyze data, have become integral to a myriad of sectors, such as finance \cite{bennett2022lead}, geology \cite{DAWSON2023105284} and ecology \cite{cazau2021multimodal,blount2022flukebook}. The depth and breadth of their analytical prowess have rendered them invaluable in deriving actionable insights from seemingly impenetrable datasets. This transformative ability has set the stage for pioneering breakthroughs in scientific pursuits, catalyzing more informed decision-making processes and enabling solutions to once-intractable real-world challenges. 

Traditional machine-learning techniques have been central to data mining for years, and researchers have emphasized interpretability as much as performance. Interpretability is paramount, not just for trust but also for validation, debugging, and understanding the broader implications of AI-driven decisions. Mi et al. \cite{9234594} classified interpretable methods into two categories: models that are inherently interpretable and those that use external mechanisms for explanation. This classification aids subsequent researchers in selecting an appropriate interpretable model for their needs. Sheykhmousa et al. \cite{sheykhmousa2020support} investigated the efficacy of machine-learning algorithms, specifically Random Forest and Support Vector Machines, for remote sensing image classification. They discussed the interpretability-accuracy tradeoff in machine learning classification algorithms and found that with increasing accuracy, these models possess decreasing interpretability. Kliegr et al. \cite{kliegr2021review} investigated the effects of cognitive biases on machine learning models' interpretation, they found that cognitive biases may affect human understanding of interpretable machine learning models, in particular of logical rules discovered from data. Researchers continue to focus on machine learning models because of their reduced computational complexity and superior interpretability when compared to deep learning models.

As the frontier of AI advances, models are being designed with sophisticated structures and concepts to achieve outstanding performance in tackling real-world challenges. These include faster detection speeds, enhanced accuracy, robustness, and superior generalization capabilities. Yet, the emphasis for many of these models remains on their end outputs across diverse tasks \cite{cao2022towards}. Zhou et al. \cite{zhou2022ladder} introduced the LADDER framework to tackle the susceptibility of Deep Neural Networks (DNNs) to adversarial attacks. By emphasizing perturbations in the latent space, directed by an SVM decision boundary with attention, LADDER strives for a balance between standard model accuracy and defense against adversarial intrusions. Extensive experiments on datasets like MNIST, SVHN, CelebA, and CIFAR-10 demonstrated its superiority over other leading baselines in terms of both accuracy and robustness. Huawei Noah’s Ark Lab \cite{guo2022cmt} introduced a transformer-CNN hybrid network that leverages transformers for capturing long-range image dependencies and CNNs (Convolutional Neural Networks) for local information extraction. This hybrid design delivers a superior balance between accuracy and efficiency compared to previous CNN and transformer models. It exhibits impressive generalization across various datasets including CIFAR10, CIFAR100, Flowers, and COCO. Nonetheless, the breakthroughs of these deep-learning-based models come at the cost of extensive computational resources. Beyond the primary research thrust of enhancing their performance, the challenge of interpretability in these powerful deep-learning architectures remains a pressing concern.

Recognizing these challenges, TabNet, or ``Attentive Interpretable Tabular Learning" \cite{Arik_Pfister_2021}, was introduced. TabNet synergizes the strengths of traditional machine learning, specifically tree-based models, with the power of deep learning techniques like transformers and the attention mechanism. Its distinct advantages include automatic feature selection, a lightweight deep-learning model design, and enhanced interpretability. However, there is always room for improvement, especially in terms of accuracy and model interpretability. Another ongoing challenge is developing standardized metrics to evaluate the interpretability of AI models, ensuring they are both effective and understood by users.

Given all of the existing challenges mentioned above, this work proposed the improved InterpreTabNet inspired by the previous TabNet, which has a better performance in classification and interpretation. Specifically, the contributions of this paper are as follows:
\begin{enumerate}
    \item Improve TabNet by integrating a Multilayer Perceptron (MLP) structure into its Attentive Transformer module. This augmentation bolsters the model's feature extraction capability, ensuring a stable foundation for explicating feature importance.
    \item Design a new Multi-branch Weighted Linear Unit activation function (Multi-branch WLU), which provides stronger model expressiveness, better gradient propagation characteristics, and higher computational stability, thereby enabling the modeling of more complex relationships and improving the model's generalization capabilities.
    \item Substitute the Sparsemax activation function with the Entmax sparse activation function. This adjustment strikes a balance between sparsity and expressive power. It plays a pivotal role in feature extraction, enhancing gradient propagation characteristics, and in deep-learning models, it acts as a deterrent to overfitting.
    \item Define a new evaluation metric, InterpreStability, which captures the stability of the model's interpretability. This metric's effectiveness has been validated, and it has been employed to analyze various models.
    \item Validate the model on different datasets from diverse scenarios.
\end{enumerate}
By implementing these modifications, the proposed InterpreTabNet model achieved significant improvements in both classification and interpretability metrics compared to the baseline models. Specifically, we observed an increase in the AUC score by 1.53\%, and a notable enhancement in the InterpreStability metric, demonstrating the model's robustness and interpretability. These findings highlight the effectiveness of our strategy in navigating the challenges currently prevalent in data analysis and deep learning model performance. Our contribution specifically directs attention towards the importance of interpretability research in deep learning models. Furthermore, we provide quantifiable tools and a reference point for future studies in this area.

The remainder of this paper is organized as follows:
\begin{enumerate}
    \item Section \ref{sec2} reviews the related work, discussing the key contributions and limitations of model interpretability assessment methods, and existing machine learning and deep learning models for tabular data mining tasks.
    \item Section \ref{sec3} introduces the datasets utilized in this paper and provides details on data collection and preprocessing.
    \item Section \ref{sec4} presents the methodology, offering an in-depth look at the architecture and functionalities of our InterpreTabNet model. It also introduces the methodology of our novel evaluation metric.
    \item Section \ref{sec5} details the experiments and results, providing a comparative analysis of InterpreTabNet against other baseline models on various datasets.
    \item Section \ref{sec6} discusses the implications, limitations, and future work.
    \item Finally, Section \ref{sec7} concludes the paper and provides a summary of the key findings.
\end{enumerate}

\section{Related Work} \label{sec2}
\subsection{Machine Learning Approaches for Tabular Data}
\subsubsection{Logistic Regression}

Logistic Regression \cite{wa2023regression} is a widely used linear classification algorithm that models the probability of an instance belonging to a particular class. The logistic function (also known as the sigmoid function) is used to transform a linear combination of input features into a probability score between 0 and 1. The logistic regression model is defined as Equation \ref{eq: logistic}.
\begin{equation}
P(y=1|\mathbf{x}) = \frac{1}{1 + e^{-(\mathbf{w} \cdot \mathbf{x} + b)}}
\label{eq: logistic}
\end{equation}
where $P(y=1|\mathbf{x})$ represents the probability of the positive class, $\mathbf{w}$ is the vector of weights, $b$ is the bias, $\mathbf{x}$ is the feature vector.

Logistic Regression is interpretable and provides coefficients for features. It works well for linearly separable data and can handle high-dimensional datasets. However, Logistic Regression assumes that the relationship between features and the log-odds of the response variable is linear and, thereby may not perform well on complex, non-linear datasets.

\subsubsection{Support Vector Machines}

Support Vector Machines (SVM) \cite{bhavsar2013intrusion} is a powerful classification algorithm that finds the hyperplane that maximizes the margin between different classes. SVM can handle both linear and non-linear classification tasks by using different kernel functions such as the linear, polynomial, or radial basis function kernel. The decision function of a linear SVM is given by Equation \ref{eq: svm}.
\begin{equation}
f(\mathbf{x}) = \mathbf{w} \cdot \mathbf{x} + b
\label{eq: svm}
\end{equation}
where $f(\mathbf{x})$ is the decision function, $\mathbf{w}$ is the vector of weights, $b$ is the bias, $\mathbf{x}$ is the feature vector.

SVM is effective in high-dimensional spaces and can handle non-linear classification tasks, addressing problems that are tricky for Logistic Regression. Meanwhile, it is less prone to overfitting due to the margin maximization. Nevertheless, SVM can be computationally expensive, especially for large datasets. And settings for the method, such as the choice of the kernel and hyperparameters, can impact the model's performance.

\subsubsection{Random Forest}
Random Forests \cite{breiman2001random} are ensemble methods, combining the predictions of numerous decision trees to improve generalization and reduce overfit. Mathematically, given a set of decision trees \( \{ T_1, T_2, \ldots, T_n \} \), the Random Forest prediction \( \hat{y} \) is given as Equation \ref{eq: randomforest}.
\begin{equation}
\hat{y} = \frac{1}{n} \sum_{i=1}^{n} T_i(x)
\label{eq: randomforest}
\end{equation}
However, Random Forest may not capture complex, non-linear relationships effectively, particularly in small datasets.

\subsubsection{LightGBM}
LightGBM (Light Gradient Boosting Machine) \cite{ke2017lightgbm}, a gradient boosting framework, employs decision trees and focuses on leaf-wise growth rather than level-wise, enabling more accurate approximations of the target variable. It uses the following objective function in Equation \ref{eq: lightgbm}.
\begin{equation}
\text{Objective} = \text{L}(y, \hat{y}) + \sum \text{Regularization}
\label{eq: lightgbm}
\end{equation}
where \( L \) is the loss function and \( \hat{y} \) is the predicted output.

LightGBM employs a leaf-wise growth strategy, reducing the complexity of splitting nodes and thus speeding up training, particularly for large-scale datasets. LightGBM uses gradient boosting, typically achieving high predictive accuracy. However, because of its leaf-wise growth, LightGBM can be sensitive to noisy data, potentially leading to overfitting.

\subsection{Deep Learning Approaches for Tabular Data}
Multilayer Perceptrons \cite{taud2018multilayer} are commonly used for tabular data. Given input \( x \), an MLP with one hidden layer can be expressed as Equation \ref{eq: mlp}.
\begin{equation}
f(x) = \sigma(W_2 \sigma(W_1 x + b_1) + b_2)
\label{eq: mlp}
\end{equation}
where \( \sigma \) is an activation function. Despite their expressive power, MLPs are often seen as black-box models, with limited interpretability.

Recurrent Neural Networks (RNNs) \cite{sherstinsky2020fundamentals} are especially useful for dealing with sequences and have been applied to tabular data with a temporal dimension. The core idea behind an RNN is to maintain a hidden state that captures information about the sequence up to the current time step. Mathematically, the hidden state \( h_t \) and the output \( y_t \) at time step \( t \) are given by Equation \ref{eq: rnn}.
\begin{align}
h_t &= \sigma(W_{hh} h_{t-1} + W_{xh} x_t + b_h) \nonumber \\
y_t &= W_{hy} h_t + b_y
\label{eq: rnn}
\end{align}
where \( \sigma \) is an activation function, \( x_t \) is the input at time \( t \), and \( W_{hh}, W_{xh}, W_{hy}, b_h, \) and \( b_y \) are the model parameters. Like MLPs, RNNs are highly expressive but also suffer from being black-box models with limited interpretability.

\subsection{Hybrid Approaches}
Recent trends in tabular data analysis have explored hybrid models that combine the robustness and interpretability of machine learning with the expressive power of deep learning. These approaches often use shallow trees or ensemble methods for feature selection or preprocessing, coupled with neural networks for capturing complex patterns in the data. Such methods attempt to leverage the strengths of both paradigms to create models that are both accurate and interpretable. While promising, these hybrid models can sometimes suffer from complexity and a tendency to overfit, especially on smaller datasets \cite{shimmei2023can,bejani2021systematic,ahmad2020fake}.

In this paper, we utilize TabNet, a model that ingeniously merges the decision-tree mechanism of traditional machine learning with the expressive power of deep learning techniques. TabNet introduces the concept of ``decision steps" that generate relatively simple and intuitive rules, thus offering a level of interpretability that is often lacking in deep neural network models. While TabNet represents a significant step toward reconciling performance and interpretability, there are still opportunities for improvement. This leads to our work where we introduce InterpreTabNet. Details on these novel contributions will be discussed in the subsequent sections.

\subsection{Evaluating Model Interpretability Stability}
Assessing the stability of model interpretability methods has become increasingly important in recent years \cite{tseng2020fourier,zafar2021deterministic,8852158}. Existing approaches aim to evaluate the consistency and robustness of feature importance \cite{10.1145/3298689.3347043} scores and explanations provided by machine learning models.

One widely used approach involves permutation-based techniques. These methods randomly shuffle or permute feature values and observe their impact on the model's predictions and associated explanations. The goal is to determine whether small data perturbations lead to significant changes in feature importance. Examples of such techniques include permutation feature importance (PFI) \cite{altmann2010permutation} and permutation SHAP (SHapley Additive exPlanations). We will use the PFI as a standard evaluation metric to validate the effectiveness of our novel metric in Section \ref{sec5}. While permutation-based methods offer valuable insights, they come with certain limitations. They can be computationally expensive, particularly for large datasets and complex models. Additionally, they may not address all aspects of stability, especially in cases involving high-dimensional data and intricate model architectures.

Despite their utility, current stability evaluation methods have notable challenges and limitations:
\begin{enumerate}
    \item Computational Cost: Permutation-based techniques can be resource-intensive, limiting their practicality for real-world applications.
    \item Scope: Some methods are tailored to specific model types or explanation approaches, restricting their applicability across diverse machine learning scenarios.
    \item Interpretability vs. Accuracy Trade-off: Certain stability evaluation methods may prioritize interpretability over predictive accuracy, which may not align with the objectives of machine learning practitioners.
\end{enumerate}

To address the challenges and limitations of existing stability evaluation techniques, we propose InterpreStability, a novel and comprehensive method. InterpreStability is designed to efficiently assess model interpretability stability across various machine learning models and explanation methods. Leveraging advanced statistical techniques and optimization algorithms, InterpreStability offers a scalable solution for practitioners seeking reliable and stable model explanations.

In the subsequent sections, we will provide a detailed overview of InterpreStability, its methodology, and empirical results showcasing its effectiveness.

\section{Materials} \label{sec3}
\subsection{Data Collection}
In this study, we employ the ``Default of Credit Card Clients'' dataset \cite{misc_default_of_credit_card_clients_350}, sourced from the University of California, Irvine's Machine Learning Repository, as one of the datasets for our analysis. Comprising 30,000 instances, the dataset captures the credit default scenarios among customers in Taiwan and is publicly available for diverse academic and commercial adaptations. The dataset consists of 25 integer-type attributes; the ``default.payment.next.month'' serves as the binary response variable, coded as 1 for default and 0 for non-default. The remaining 24 attributes, enumerated in Table \ref{tab: creditcard}, function as explanatory variables.

\begin{table}[h]
\caption{Display of the detailed information of explanatory variables in the dataset}\label{tab: creditcard}
\begin{tabularx}{\textwidth}{@{}lX@{}}
\toprule
Attribute Name & Description  \\
\midrule
ID    & ID number of credit card clients  \\
LIMIT\_BAL  & Amount of the given credit (Taiwan New Dollar)  \\
SEX    & Gender of credit card clients: male=1, female=2  \\
EDUCATION\footnotemark[1]  & Education degree: graduate school=1, university=2, high school=3, others=4  \\
MARRIAGE\footnotemark[1] & Marital status: married=1, single=2, others=3 \\
AGE & Age of clients  \\
PAY\_0, PAY\_2$\sim$PAY\_6\footnotemark[1] & Past monthly payment records from April to September 2005: pay duly=-1, payment delay for one month=1, payment delay for two months=2,..., payment delay for nine months and above=9. \\
BILL\_AMT1$\sim$BILL\_AMT6 & Amount of bill statement (Taiwan New Dollar) from April to September 2005  \\
PAY\_AMT1$\sim$PAY\_AMT6 & Amount of previous payment (Taiwan New Dollar) from April to September 2005  \\
\botrule
\end{tabularx}
\footnotetext[1]{These variables' distribution could be modified and improved, which will be discussed in the following Section \ref{sec: preprocessing}.}
\end{table}

In addition to the ``Default of Credit Card Clients" dataset, we also incorporate three additional small-volume datasets for validation in Sections \ref{sec5} and \ref{sec6}, namely Iris, Breast Cancer Wisconsin, and Digits. (1) The Iris dataset, sourced from the UCI Machine Learning Repository \cite{misc_iris_53}, consists of 150 samples of iris flowers, with each sample belonging to one of three species. It contains four features, namely sepal length, sepal width, petal length, and petal width, all of which are continuous numerical values. This dataset serves as a classic benchmark for multi-class classification tasks. (2) The Breast Cancer Wisconsin dataset, also obtained from the UCI Machine Learning Repository \cite{misc_breast_cancer_wisconsin_(diagnostic)_17}, comprises 569 instances of breast cancer biopsies, classified as either malignant (cancerous) or benign (non-cancerous). It encompasses 30 real-valued features characterizing cell nuclei properties extracted from breast cancer biopsies. (3) The Digits dataset, available as part of the Scikit-learn (Sklearn) library \cite{scikit-learn}, consists of $8 \times 8$ pixel images of handwritten digits (0-9). It totals 1,797 samples, which can be obtained as flattened image numerical data for digit classification by Sklearn. All three datasets are devoid of missing values and require no preprocessing, rendering them ideal and standard candidates for validation experiments.

\subsection{Data Preprocessing} \label{sec: preprocessing}
\subsubsection{Data Clean}
To prepare the ``Default of Credit Card Clients'' dataset for classification tasks---specifically, to predict whether a client will default on a payment---we undertook a comprehensive data cleaning and analysis process. Our statistical examination revealed anomalies in three specific attribute categories:

\begin{enumerate}
    \item The EDUCATION attribute is expected to contain integer values ranging from 1 to 4. However, we observed extraneous values such as 0, 5, and 6. To standardize this feature, we recoded these anomalies to 4, categorizing them as ``others.''
    \item Similarly, the MARRIAGE attribute contained an unexpected value of 0. To bring consistency to the dataset, this value was recoded to 3, also falling under the ``others'' category.
    \item In the series of PAY attributes, we noted an inconsistency in naming, starting with PAY\_0 but skipping PAY\_1. To achieve a coherent naming convention, we renamed the PAY\_0 attribute to PAY\_1. Furthermore, we identified a peculiar value of -2 in these attributes, which we standardized to -1 to denote the ``pay duly'' category.
\end{enumerate}

These modifications aim to improve the consistency and interpretability of the dataset, thereby enhancing the reliability of the subsequent classification models.

\subsubsection{Data Splitting}
To prepare the data for both training and evaluation, we partitioned the dataset into training, validation, and test sets following a 7:1:2 ratio. Specifically, the training set contains 21,000 samples, the validation set has 3,000, and the test set comprises 6,000 samples. It is worth noting that the stratified splitting technique was applied to ensure that the distribution of positive and negative samples in each subset mirrors the original dataset. 

The dataset in this study exhibits class imbalance, containing 6,636 positive samples as opposed to 23,364 negative samples, leading to a ratio of roughly 1:4. Initial efforts to mitigate this imbalance, such as employing the Synthetic Minority Over-sampling Technique (SMOTE) \cite{chawla2002smote} and auto-adjusting class weights during training, unfortunately led to a slight decrease in model performance. It is crucial to recognize that not all machine learning algorithms stand to benefit from oversampling techniques like SMOTE. In fact, ensemble methods like Random Forests and Gradient Boosting Trees generally exhibit better robustness when dealing with imbalanced datasets. Additionally, it is worth noting that in the financial domain, data balancing techniques may introduce specific drawbacks, such as increasing the likelihood of Type \uppercase\expandafter{\romannumeral1} errors (false positives), which could lead to unnecessary or misguided interventions.

\section{Methods}\label{sec4}
\subsection{Overview of the TabNet Architecture}
\begin{figure}[h]
  \centering
  \includegraphics[width=1\textwidth]{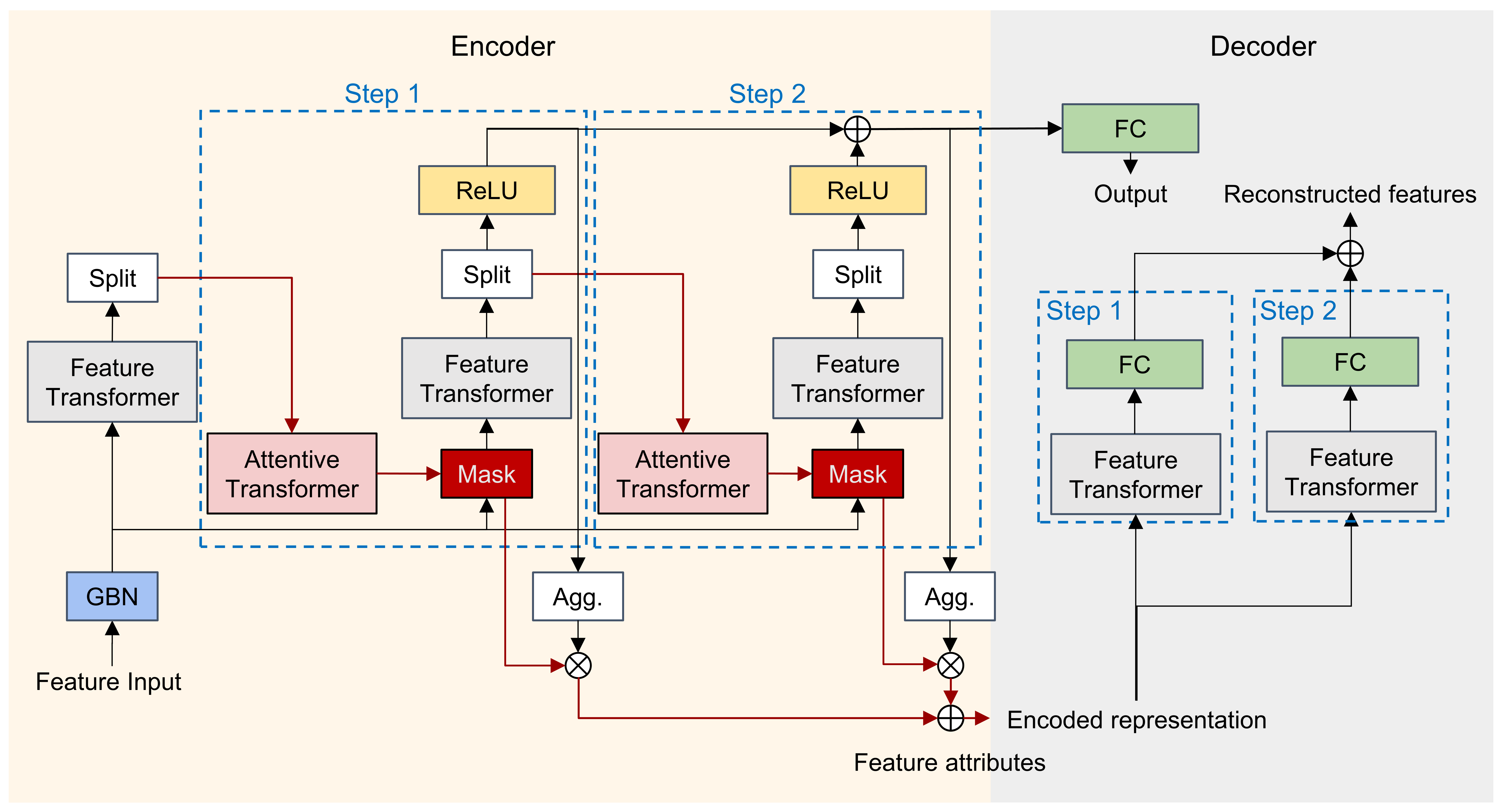}
  \caption{The architecture of TabNet's encoder and decoder. The decision steps of TabNet can be designed according to particular requirements, and this figure shows a two-step architecture.}
  \label{fig: tabnet}
\end{figure}

TabNet is a deep learning architecture designed specifically for tabular data classification and regression tasks. Unlike conventional architectures that primarily focus on image or text data, TabNet excels at capturing the intricacies of structured data in tabular form. In this subsection, we provide an in-depth explanation of the various design features and architectural components that constitute the TabNet model. The overall architecture of TabNet's encoder and decoder is displayed in Figure \ref{fig: tabnet}.

\begin{enumerate}
    \item \textit{Input layer}:
    
    The input layer of TabNet takes in tabular data, typically in the form of a matrix where rows represent instances and columns represent features. The input data is then normalized to prepare it for the subsequent layers.

    \item \textit{Decision step and Feature Transformer blocks:}
    
    As depicted in Figure \ref{fig: tabnet}, TabNet is structured into a series of decision steps, each housing one or more Feature Transformer blocks. The Feature Transformer serves as the cornerstone of TabNet's architecture, playing a crucial role in learning intricate relationships among various input features. It functions as an advanced feature selector that dynamically focuses the model's attention on salient variables during each decision step.
    
    A typical 4-layer Feature Transformer block, as shown in Figure \ref{fig: featuretransformer}, has two layers that are shared across all decision steps and two that are decision step-specific. Each layer is made up of three primary components: (1) a Fully Connected (FC) layer \cite{ma2017equivalence} that starts the transformation by capturing linear relationships between input features, followed by (2) a Ghost Batch Normalization (GBN) \cite{hoffer2017train} that standardizes the output of the FC layer. GBN is a modified form of Batch Normalization designed to work well with mini-batch gradient descent methods. By dividing the mini-batch into smaller ``ghost" batches, GBN provides a balance between the statistical efficiency of Batch Normalization and the computational efficiency of smaller batch sizes. GBN stabilizes the learning process and minimizes the risk of overfitting, especially when training data are limited. (3) Finally, a Gated Linear Unit (GLU) \cite{dauphin2017language} introduces non-linearities that enable the dynamic feature selection. The shared layers help to economize on the model's parameters, while the decision step-specific layers enable TabNet to focus adaptively on the most important features at each decision step.

    \begin{figure}[h]
      \centering
      \includegraphics[width=1\textwidth]{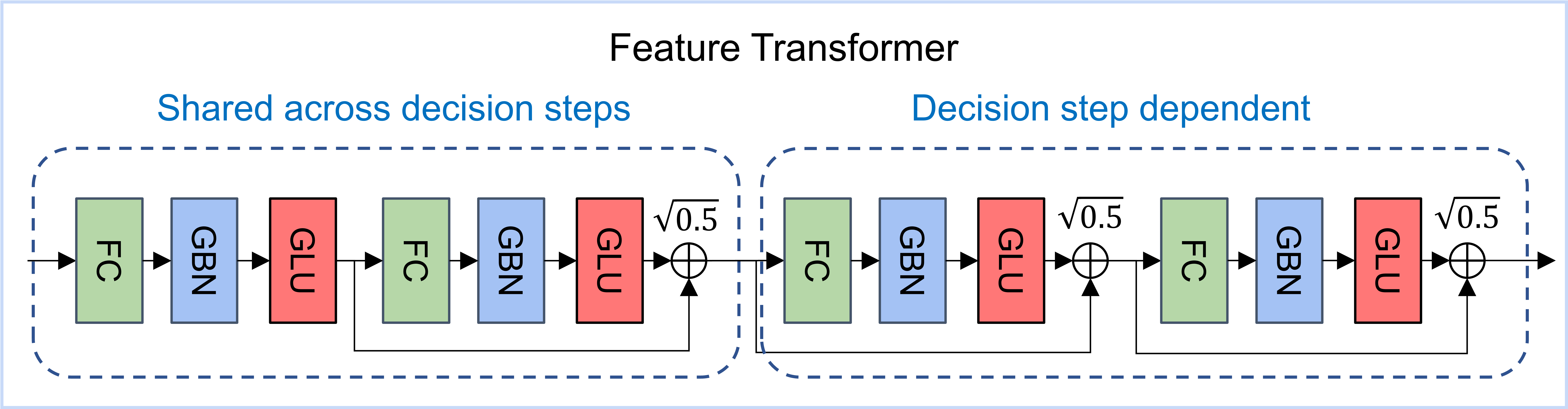}
      \caption{The architecture of 4-layer Feature Transformer}
      \label{fig: featuretransformer}
    \end{figure}

    \item \textit{Sparsemax-Attentive Transformer:}
    
    The Attentive Transformer adaptively learns complex patterns and relationships in the data, responsible for learning to pay attention to the most relevant features for the task at hand. Its primary constituents are an FC layer, GBN, and a Sparsemax activation function. Figure \ref{fig: attentivetransformer} elaborates
    its architecture.
    
    \begin{figure}[h]
      \centering
      \includegraphics[width=0.5\textwidth]{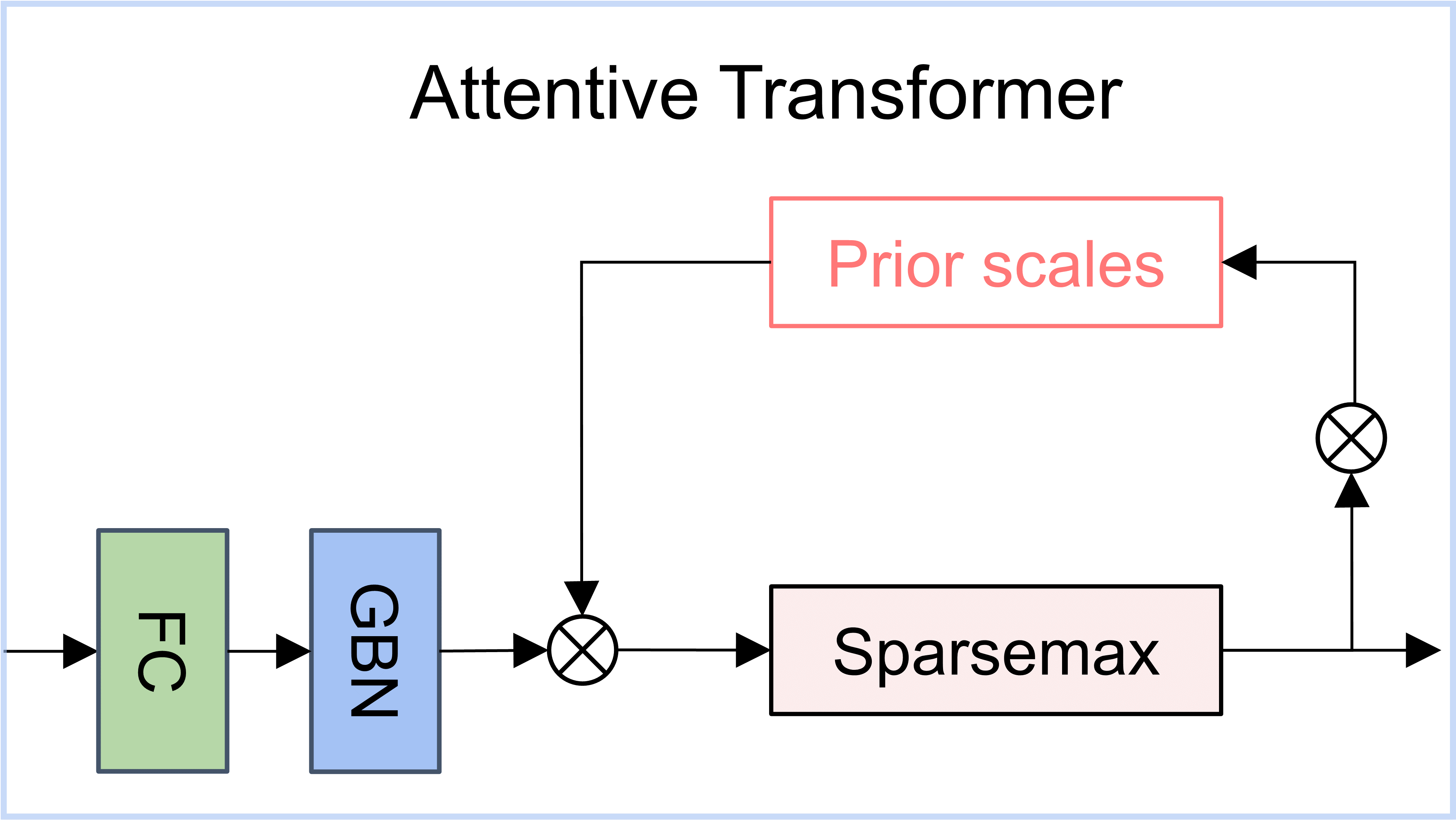}
      \caption{The architecture of Attentive Transformer}
      \label{fig: attentivetransformer}
    \end{figure}
    
    The FC layer captures intricate relationships between different features, taking as input the features selected by the Masking process and generating a dense representation. Following the FC layer, GBN is applied to stabilize the network and assist in generalization. The output of GBN is aggregated with the amount of feature that has been selected before the current step and then delivered to the final layer, the Sparsemax activation function. The Sparsemax activation function \cite{martins2016softmax} is a generalization of the traditional Softmax function \cite{jang2016categorical} but produces a sparse probability distribution over the features. The sparsity induced by Sparsemax enables the model to focus on a smaller subset of important features, thereby making the model's decision-making process more interpretable.
    
    The output of the Sparsemax function is used to create a mask that selects the most important features for the subsequent decision step. This mask is applied element-wise to the original feature vector, emphasizing relevant features while attenuating less important ones. By iteratively refining these masks through back-propagation, the Attentive Transformer guides the model to focus on features that are crucial for the task.

    \item \textit{Encoder's outputs:}
     
    The outputs from various decision steps are aggregated and these encoded representations are inputs for the decoder module. This aggregation enhances the model's ability to capture complex relations and dependencies among the features.

    \item \textit{Decoder:}
    
    The decoder processes the encoded features through multiple decoding steps to obtain reconstructed features and final output in terms of specific tasks. Each decoding step consists of two core components: a Feature Transformer and an FC layer. The Feature Transformer functions in a manner similar to its counterpart in the encoding phase, transforming the input features at each decoding step. The outputs from each decoding step are then aggregated to produce the final set of reconstructed features, serving as an approximation of the original input features. Additionally, depending on the task, an appropriate activation function is applied to the output of the final FC layer. For instance, in classification tasks, a Softmax function is typically used to convert the output into probability scores over the different classes. For regression tasks, no activation function or a linear activation function might be applied. This comprehensive mechanism enables the model not only to generate robust and effective feature representations but also to provide insights into the reconstruction process, thereby aiding in model interpretability and refinement.

\end{enumerate}

By thoughtfully integrating these elements, TabNet provides a powerful yet interpretable framework for tabular data modeling. In the subsequent sections, we introduce our advancements and customizations that further enhance the capabilities of the standard TabNet architecture.

\subsection{Architecture and Innovations in InterpreTabNet}

\begin{figure}[h]
      \centering
      \includegraphics[width=1\textwidth]{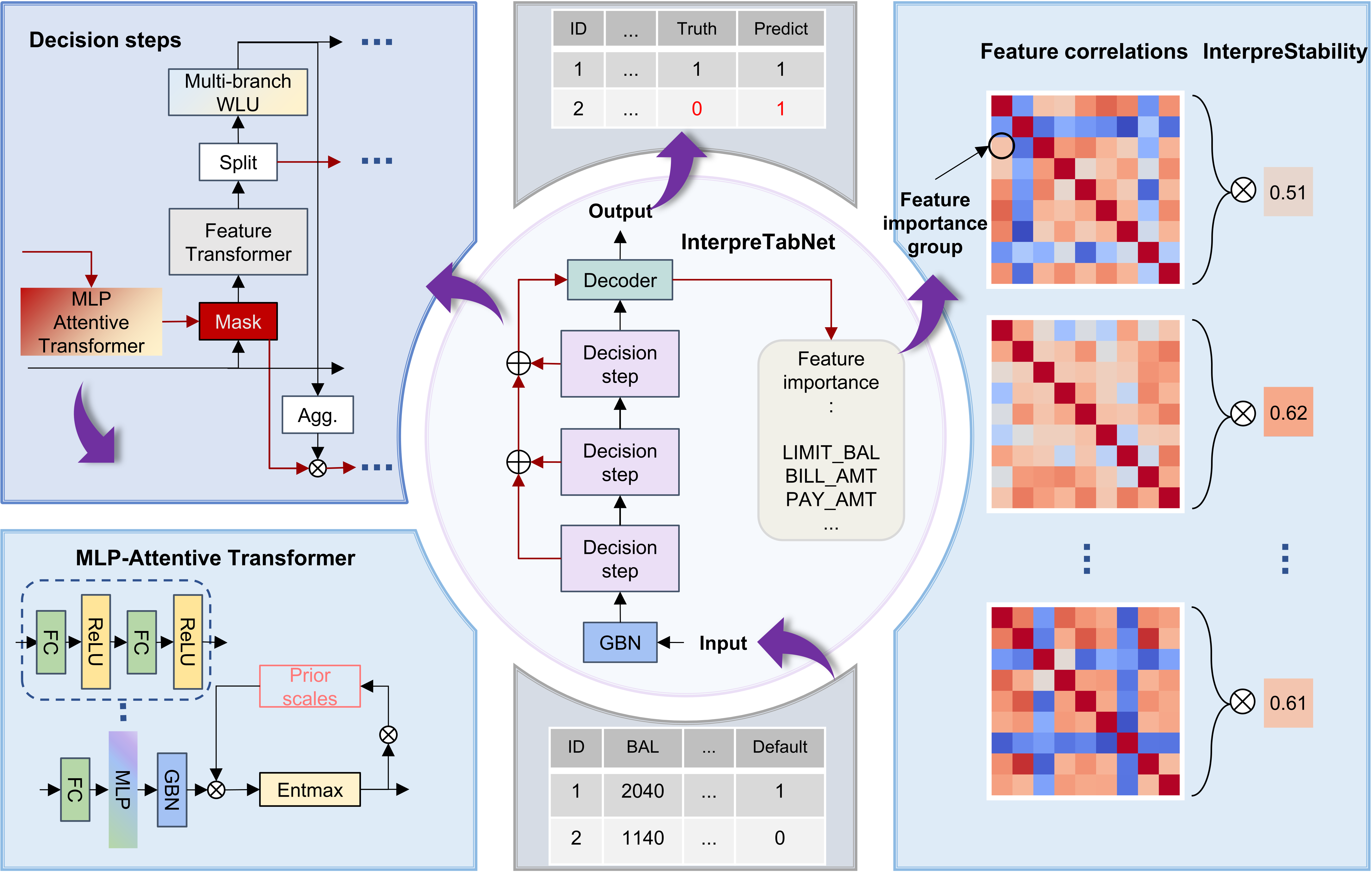}
      \caption{The architecture of InterpreTabNet and primary innovations in this work. (1) The center part is the main framework of InterpreTabNet, consisting of three decision steps and a decoder with three decoding steps. Each output of one decision step is aggregated with the following one and ultimately passed to the decoder. The decoder gives the final output, selected features, and feature importances. (2) At the top left corner is each decision step in the encoder, where we propose the Multilayer Perceptron-Attentive Transformer (MLP-Attentive Transformer) and Multi-branch WLU. (3) The bottom left corner is the detailed architecture of the MLP-Attentive Transformer. (4) The entire right part visualizes the scheme of our novel evaluation metric, InterpreStability. Heatmaps represent feature correlations, whose elements are groups of feature importance. The little squares denote specific InterpreStability values after the calculation. (5) The middle bottom and top are the input and output of this paper's classification task, respectively. The input is the tabular data, while the output is the classification result.}
      \label{fig: interpretabnet}
    \end{figure}

Building upon the foundational architecture of TabNet, we introduce InterpreTabNet, a refined and enhanced model designed to address some of the limitations and challenges associated with the original implementation. A schematic overview of the InterpreTabNet framework is provided in Figure \ref{fig: interpretabnet}. In the following sections, we will elaborate on the specific improvements that have been implemented, detailing how each contributes to increased model performance, robustness, and interpretability. These innovations aim to push the boundaries of what is possible in the realm of interpretable machine learning for tabular data.

\subsubsection{Multi-branch Weighted Linear Unit}

Activation functions play a crucial role in the learning capabilities of neural networks. The original TabNet employs ReLU (Rectified Linear Unit) \cite{nair2010rectified} as the activation function in its encoder's decision steps. However, this choice has several shortcomings, such as the dying ReLU problem, where neurons can sometimes get stuck during training and stop updating their weights, leading to a loss of model capacity. This drawback derives from the calculation of ReLU, as expressed in Equation \ref{eq: relu}, because it would set some values as zero.
\begin{equation}
        f(x) = \max(0, x)
        \label{eq: relu}
    \end{equation}

The Parametric Rectified Linear Unit (PReLU) \cite{7410480} generalizes ReLU by introducing a learnable parameter \( \alpha \) rather than a fixed parameter like Leaky RelU \cite{maas2013rectifier}. The learnable parameter scales the output for negative input values and offers a balance between ReLU and a linear function, allowing the network to learn the most suitable activation behavior during training. It is represented by Equation \ref{eq: prelu}.
\begin{equation}
    f(x) = \max(0, x) + \alpha \min(0, x)
    \label{eq: prelu}
\end{equation}

The Exponential Linear Unit (ELU) \cite{clevert2015fast} was developed to address some of the deficiencies observed in both ReLU and its generalization, PReLU. Unlike ReLU, ELU can produce negative outputs via the term \( \alpha(e^x - 1) \), thereby mitigating the dying ReLU problem. This characteristic introduces non-zero gradients for negative inputs, ensuring a smoother optimization landscape compared to ReLU. Moreover, the exponential component allows ELU to push the mean activation closer to zero, which speeds up the learning process. In contrast to PReLU, which introduces learnable parameters to adjust the slope for negative inputs, ELU uses a fixed exponential function for this purpose, effectively introducing non-linearity without the need for additional parameters to be learned. The ELU function is defined as Equation \ref{eq: elu}.
\begin{equation}
    f(x) = \begin{cases} 
    x & \text{if } x > 0 \\
    \alpha(e^x - 1) & \text{otherwise}
    \end{cases}
    \label{eq: elu}
\end{equation}
Here, \( \alpha \) is a hyperparameter, and the term \( \alpha(e^x - 1) \) ensures a smoother and more balanced behavior, improving the generalization performance of the model.

Lastly, the Sigmoid Linear Unit (SiLU) \cite{ramachandran2017searching}, also known as the Swish function, is another variant that attempts to offer a balance between linear and non-linear behavior. SiLU offers smoothness across its domain, allowing the model to take both negative and positive values into account. This smoothness property leads to easier optimization and potentially better generalization. The function is represented by Equation \ref{eq: silu}.
\begin{equation}
    f(x) = x \cdot \frac{1}{1 + e^{-x}}
    \label{eq: silu}
\end{equation}

Each of these activation functions offers specific advantages over simple ReLU, such as mitigating the dying ReLU problem, introducing learnable parameters, or offering smoother gradients, thus enhancing the training and generalization capabilities of neural networks. Inspired by these merits, we propose a Multi-branch Weighted Linear Unit by integrating ELU, PReLU, and SiLU. Each branch corresponds to one of these activation functions, and they are combined with learnable weights as expressed by Equation \ref{eq: wlu}.
\begin{equation}
    f(x) = \alpha \times \text{ELU}(x) + \beta \times \text{PReLU}(x) + \gamma \times \text{SiLU}(x)
    \label{eq: wlu}
\end{equation}
where \( \alpha, \beta, \gamma \) are weight parameters initially set as 0.6, 0.2, and 0.2 according to the experimental results.

The Multi-branch Weighted Linear Unit replaces the ReLU layer of the encoder's each decision step as shown in Figure \ref{fig: interpretabnet}. Our experimental results confirm that this new activation function significantly improves the performance of TabNet.

\subsubsection{Multilayer Perceptron-Attentive Transformer}
\label{sec: MLAT}
As illustrated in the left part of Figure \ref{fig: interpretabnet}, we introduce a Multilayer Perceptron-Attentive Transformer module in each decision step of the encoder. This enhancement aims to improve the feature extraction capabilities of the original Attentive Transformer.

The MLP-Attentive Transformer module is situated between the FC and the GBN layers of the conventional Attentive Transformer. The MLP module consists of two sequences of FC and ReLU layers, represented as Equation \ref{eq: mlpAF}.
\begin{equation}
    \text{MLP}(x) = \text{ReLU}(\mathrm{FC}_{2}(\text{ReLU}(\mathrm{FC}_{1}(x))))
    \label{eq: mlpAF}
\end{equation}

The incorporation of the MLP module aims to increase the model's expressiveness by introducing an additional level of non-linearity. This is particularly beneficial for capturing more complex relationships between features, thus potentially leading to more accurate decision-making processes within the encoder. The combination of FC, ReLU, and GBN in the MLP-Attentive Transformer module also ensures better gradient flow and more stable training, especially when dealing with high-dimensional data or deep architectures.

\subsubsection{Entmax Activation Function}

In the original design of the Attentive Transformer, Sparsemax is utilized as the activation function, which is defined as Equation \ref{eq: sparsemax}.
\begin{equation}
    \text{Sparsemax}(z)_i = \max\left(0, z_i - \tau(z)\right),
    \label{eq: sparsemax}
\end{equation}
where \( \tau(z) \) is a threshold function that ensures the sum of the resulting vector is 1.

However, we replace Sparsemax with Entmax \cite{peters2019sparse} to enhance the model's performance, which is defined as Equation \ref{eq: entmax}.
\begin{equation}
    \text{Entmax}(z)_i = \text{argmin}_{s} \left( \|s - z\|_2^2 - \alpha H(s) \right),
    \label{eq: entmax}
\end{equation}
where \( H(s) \) is the entropy of the vector \( s \) and \( \alpha \) is a hyperparameter controlling the sparsity of the output.

The primary reason for replacing Sparsemax with Entmax is to introduce a level of sparsity controlled by the hyperparameter \( \alpha \). This provides a more flexible activation function, allowing for a better compromise between sparsity and expressiveness, thereby enhancing the model's ability to focus on critical features.

\subsection{The Concept and Calculation of InterpreStability}

One of the prominent features of TabNet is its inherent capability to provide feature importance scores, which are crucial for interpretability. The feature importance in TabNet is primarily derived from the attention masks generated at each decision step. 

Although TabNet presents a novel, effective, and authoritative mechanism for feature importance calculation, there exists no standard metric for evaluating the selection process and performance of these calculated features. This issue is not unique to TabNet; the evaluation of interpretability in explainable artificial intelligence (XAI) models has been a topic of ongoing research, and various metrics and frameworks have been proposed to evaluate the interpretability of models \cite{tjoa2020survey}. However, there is not a universally agreed-upon measure or standard for interpretability, partly because the concept itself can be domain-specific and user-specific. In light of this, we introduce the InterpreStability metric, which combines feature importance scores to assess the interpretive stability of XAI models.

\subsubsection{Feature Importance in TabNet}

At each decision step $t$, an attention mask $M_t$ is generated by the Attentive Transformer as Equation \ref{eq: mask}.
\begin{equation}
    M_t = \text{Entmax}( \text{GBN}(\text{MLP}(\text{FC}(Z_{t-1}))))
    \label{eq: mask}
\end{equation}
Here, FC denotes a Fully Connected layer, MLP represents the module mentioned in Section \ref{sec: MLAT}, GBN denotes Ghost Batch Normalization, and Entmax is the activation function. $Z_{t-1}$ is the output of the previous decision step, with $Z_0$ being the initial set of features.

The importance of a feature $i$ can be calculated by aggregating the attention mask values across all decision steps and potentially across multiple data samples. The calculation of a feature's importance is expressed in Equation \ref{eq: featureimportance}.
\begin{equation}
    \text{Feature importance}(i) = \sum_{t=1}^{T} \sum_{n=1}^{N} M_t^{(n,i)}
    \label{eq: featureimportance}
\end{equation}
where $T$ is the total number of decision steps, $N$ is the number of samples, and $M_t^{(n,i)}$ is the mask value for feature $i$ at decision step $t$ for sample $n$.

Features that frequently have higher mask values are deemed more important for the model's decision-making process. This built-in feature importance mechanism lends TabNet the capability for high interpretability.

\subsubsection{InterpreStability}

Although most explainable models, much like LIME (Local Interpretable Model-Agnostic Explanations) \cite{zafar2021deterministic} and SHAP, provide a mechanism for computing feature importance at each decision step, they lack a suitable metric to evaluate the performance of these computations, such as stability. To address this, our paper introduces the InterpreStability metric, based on the InterpreTabNet model. This metric calculates the feature importance for different random sub-samples of the dataset and then evaluates the Pearson correlation between these sets to generate a correlation matrix. The final InterpreStability score is a weighted sum of all these correlation coefficients.

Figure \ref{fig: algorithm} gives an overview of InterpreStability's methodology, and the detailed methodology of InterpreStability is shown below:

\begin{figure}[h]
      \centering
      \includegraphics[width=1\textwidth]{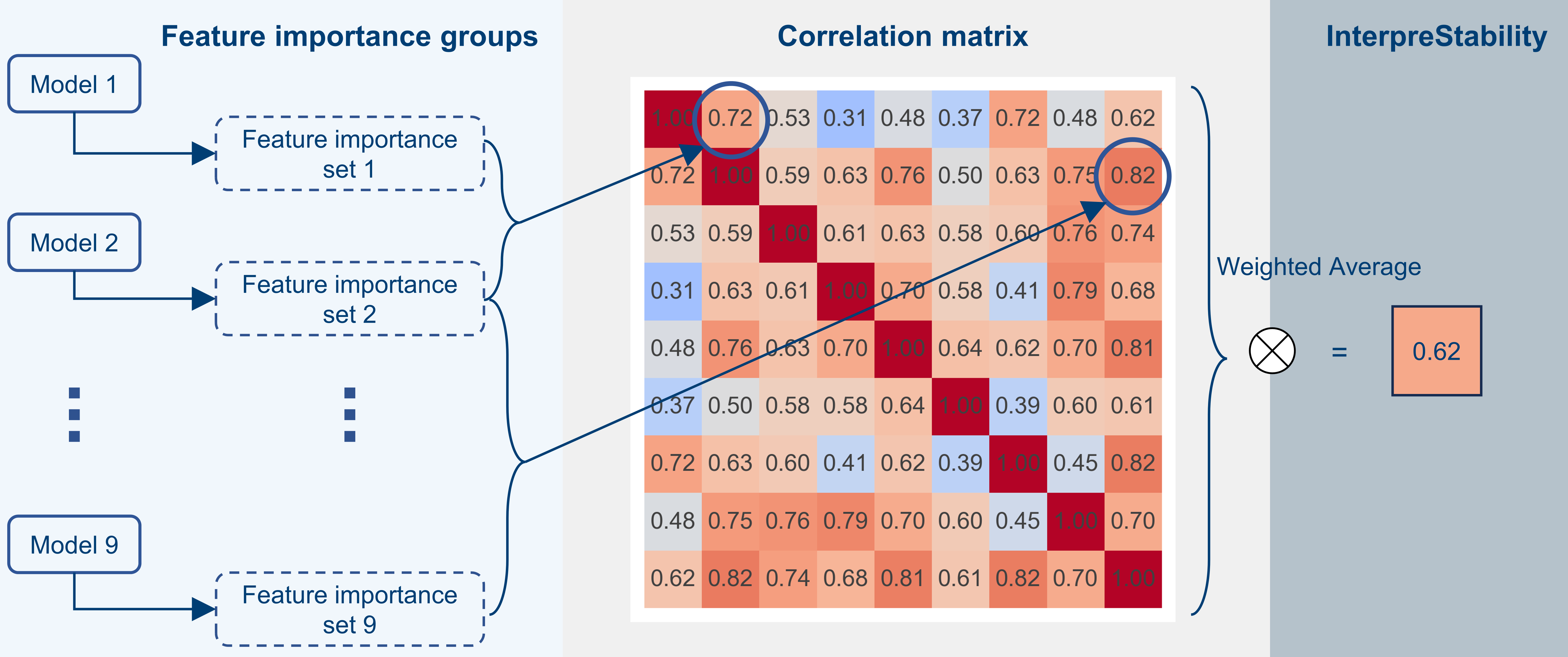}
      \caption{The workflow of InterpreStability's methodology. (1) Each model trains on different subsets or the entire dataset but is split and read in random order, then these models generate a particular feature importance set. (2) Pearson correlation matrix is calculated by using each pair of two feature importance sets. (3) The InterpreStability value could be attained by computing weighted-average correlation coefficients.}
      \label{fig: algorithm}
\end{figure}

\begin{enumerate}
    \item \textit{Compute feature importance for subsets}:
    
    Let \( X_i \) be a random subset of the dataset (also could be the whole dataset that split and read in random order), and the feature importance \( F(X_i) \) for this subset is computed using Equation \ref{eq: featureimportance} and \ref{eq: F}.
    \begin{equation}
        F(X_i) = \text{Feature importance}(X_i)
        \label{eq: F}
    \end{equation}

   \item \textit{Calculate Pearson correlation between feature importance groups}:

    The Pearson correlation \cite{cohen2009pearson} \( \rho_{ij} \) between two feature importance sets \( F(X_i) \) and \( F(X_j) \) is given by Equation \ref{eq: pearson}.
    \begin{equation}
        \rho_{ij} = \text{Pearson}(F(X_i), F(X_j))
        \label{eq: pearson}
    \end{equation}
    
    Pearson correlation coefficients are interpreted as follows:
    \begin{itemize}
        \item \( \rho \in [0.9, 1] \): Very high correlation
        \item \( \rho \in [0.7, 0.9] \): High correlation
        \item \( \rho \in [0.5, 0.7] \): Moderate correlation
        \item \( \rho \in [0.3, 0.5] \): Low correlation
        \item \( \rho \in [0, 0.3] \): Little if any correlation
        \item \( \rho < 0 \) or \( \rho > 1 \): Odd behavior, often treated as anomalies
    \end{itemize}
    
    In this study, we are more concerned with the stability of feature importance rather than its consistency across different subsets (i.e., whether a given feature is always positively or negatively important). In real-world scenarios, data distributions can vary greatly depending on the size and nature of the dataset. For instance, in smaller, region-specific datasets, the data distribution might be denser, leading the model to favor features that are strong in those specific conditions. On the other hand, in larger datasets, the feature distribution tends to be more balanced, resulting in a more balanced feature selection by the model. Investigating the consistency of feature importance across such varied conditions may not be universally applicable. In contrast, the stability of feature importance is a more generalizable attribute and offers a robust measure of the model's feature selection performance. Therefore, this study focuses on the absolute values of the Pearson coefficients for assessing stability, treating any non-positive values as indications of odd behavior in the model.

    \item \textit{Compute InterpreStability score}:

    The final InterpreStability score is a weighted sum of all the elements in the correlation matrix \( C \), derived from Equation \ref{eq: interprestability}.
    \begin{equation}
        \text{InterpreStability} = \sum_{i,j} w_{ij} \rho_{ij}
        \label{eq: interprestability}
    \end{equation}
    
    Given the interpretations of Pearson correlation coefficients, weights \( w_{ij} \) are assigned as follows:
    \begin{itemize}
        \item Very high correlation: \( w_{ij} = 1.0 \)
        \item High correlation: \( w_{ij} = 0.8 \)
        \item Moderate correlation: \( w_{ij} = 0.6 \)
        \item Low correlation: \( w_{ij} = 0.4 \)
        \item Little if any correlation: \( w_{ij} = 0.2 \)
        \item Odd behavior: \( w_{ij} = 0 \)
    \end{itemize}

    If all elements in the correlation matrix are greater than 0.9, indicating very high correlation, we will directly calculate the mean of all elements as the InterpreStability value. Otherwise, the InterpreStability value will be calculated as 1 using the following formula. Then, each interval of InterpreStability represents a different stability of the model's interpretability:
    \begin{itemize}
        \item \( \text{InterpreStability} \in [0.9, 1] \): Very high stability
        \item \( \text{InterpreStability} \in [0.7, 0.9] \): High stability
        \item \( \text{InterpreStability} \in [0.5, 0.7] \): Moderate stability
        \item \( \text{InterpreStability} \in [0.3, 0.5] \): Low stability
        \item \( \text{InterpreStability} \in [0, 0.3] \): Little if any stability
    \end{itemize}

\end{enumerate}

The InterpreStability metric provides a quantitative measure to evaluate the consistency and stability of the feature importance calculations across different subsets of the data. This adds a layer of trust and credibility to the model's interpretability, thereby enabling better decision-making processes.

\section{Experiments and Results} \label{sec5}
\subsection{Experiment Environment}

All experiments were conducted on a computer with an Intel(R) Core(TM) i7-9750H CPU, 8 GB of RAM, and an NVIDIA GeForce GTX 1650 GPU. The software environment includes Windows 11 as the operating system, Python 3.8.13 for programming, Visual Studio Code 1.81.1 as the Integrated Development Environment, and PyTorch 1.12.1 as the deep learning framework. Additional libraries such as PyTorch-TabNet 4.1.0, NumPy 1.23.1, scikit-learn 1.1.3, and matplotlib 3.6.2 were also used. To ensure reproducibility and to minimize the effects of stochasticity, experiments were run with a fixed random seed.

\subsection{Evaluation Metrics}

One of the primary evaluation metrics used in this study is the Area Under the Receiver Operating Characteristic Curve (AUC-ROC), commonly abbreviated as AUC. The ROC curve is a graphical representation that displays the true positive rate (TPR) against the false positive rate (FPR) at various decision thresholds.

Equation \ref{eq: tpr} is used for calculating the TPR.
\begin{equation}
    \text{TPR} = \frac{\text{TP}}{\text{TP} + \text{FN}}
    \label{eq: tpr}
\end{equation}
where TP represents True Positives and FN represents False Negatives. 

The FPR is calculated using Equation \ref{eq: fpr}.
\begin{equation}
    \text{FPR} = \frac{\text{FP}}{\text{FP} + \text{TN}}
    \label{eq: fpr}
\end{equation}
where FP signifies False Positives and TN represents True Negatives. 

The AUC value is the integral of the ROC curve and is computed using Equation \ref{eq: auc}.
\begin{equation}
    \text{AUC} = \int_{0}^{1} \text{TPR}(t) \, dt
    \label{eq: auc}
\end{equation}
Here, \( t \) ranges from 0 to 1 and represents the decision threshold. The AUC is a measure of the classifier's ability to distinguish between positive and negative classes. An AUC value of 0.5 suggests no discrimination, 1 signifies perfect discrimination, and 0 suggests the model is making all wrong predictions.

The AUC metric is widely regarded as a robust performance measure, especially for imbalanced datasets, and it takes into account both the sensitivity (TPR) and specificity (1-FPR) of the classifier.

In addition to the AUC-ROC metric, we also utilize Accuracy as an evaluation measure for assessing the classification accuracy of our models. Accuracy represents the proportion of correctly predicted instances out of the total number of instances in the dataset.

The accuracy of a classification model can be calculated using Equation \ref{eq: accuracy}:
\begin{equation}
\text{Accuracy} = \frac{\text{TP} + \text{TN}}{\text{TP} + \text{TN} + \text{FP} + \text{FN}}
\label{eq: accuracy}
\end{equation}

Accuracy is a fundamental metric that provides an overall view of a model's classification performance. It measures the proportion of correctly classified samples in both the positive and negative classes. An accuracy value of 1 indicates a perfect classification, while a value of 0 signifies complete misclassification.

\subsection{Training Strategy}

In our experiments, we use a two-phase approach for training TabNet: pretraining followed by fine-tuning. We make use of the PyTorch-TabNet implementation for both phases. 

We initially train an unsupervised version of the TabNet model, where the objective is to reconstruct the input. The pretraining is performed using Stochastic Gradient Descent (SGD) with an initial learning rate of \(1 \times 10^{-1}\), momentum \(0.938\), and weight decay \(1 \times 10^{-4}\). We use a batch size of \(1024\) and a virtual batch size of \(128\). The pretraining phase lasts for a maximum of \(100\) epochs, with early stopping if the validation loss does not improve for \(20\) consecutive epochs.

After pretraining, we fine-tuned the model using labeled data. We employ the same optimizer settings as during pretraining and additionally use a learning rate scheduler (ReduceLROnPlateau) that reduces the learning rate if the validation loss plateaus. The fine-tuning phase is also run for \(100\) epochs with the same early stopping criteria as the pretraining phase. Specific hyperparameter settings are shown in Table \ref{tab: hyperparameters}.

\begin{table}[h]
    \begin{tabular}{cc}
        \toprule
        Hyperparameter & Value \\
        \midrule
        Optimizer & SGD \\
        Initial Learning Rate & \(1 \times 10^{-1}\) \\
        Momentum & \(0.938\) \\
        Weight Decay & \(1 \times 10^{-4}\) \\
        Batch Size & \(1024\) (Pretraining), \(64\) (Fine-tuning) \\
        Virtual Batch Size & \(128\) (Pretraining), \(32\) (Fine-tuning) \\
        Max Epochs & \(100\) \\
        Early Stopping Patience & \(20\) \\
        Scheduler & ReduceLROnPlateau \\
        Scheduler Factor & \(0.1\) \\
        Scheduler Patience & \(5\) \\
        \bottomrule
    \end{tabular}
    \caption{Hyperparameters used for TabNet training}
    \label{tab: hyperparameters}
\end{table}

\subsection{Ablation Experiment based on Activation Functions}

Table \ref{tab: activation} presents the results of our ablation experiment, where we investigate the impact of different activation functions on the performance of the TabNet model. Activation functions play a crucial role in the model's decision-making process during training. In this experiment, we explore various activation strategies within the encoder's decision steps and evaluate their effects on Accuracy and AUC.

\begin{table}[h]
    \begin{tabular}{ccc}
        \toprule
        Activation function & Accuracy & AUC  \\
        \midrule
        ReLU & 0.7948 & 0.7681  \\
        ELU & 0.8133 & 0.7633  \\
        PReLU & 0.8218 & 0.7594  \\
        SiLU & 0.8190 & 0.7762  \\
        Multi-branch WLU-118\footnotemark[1] & 0.8193 & 0.7728  \\
        Multi-branch WLU-226\footnotemark[1] & 0.8182 & 0.7706  \\
        Multi-branch WLU-325\footnotemark[1] & 0.8183 & 0.7686  \\
        InterpreTabNet\footnotemark[2] & \textbf{0.8252} & \textbf{0.7814} \\
        \bottomrule
    \end{tabular}
    \caption{Performance of different activation functions based on the TabNet. The highest values are bold}
    \label{tab: activation}
    \footnotetext[1]{The suffixes following the three activation functions represent the weights of Multi-branch WLU. For example, ``118'' indicates the weights of ELU, PReLU and SiLU are 0.1, 0.1 and 0.8, respectively.}
     \footnotetext[2]{Our model with the weight allocation in 0.6, 0.2 and 0.2 of ELU, PReLU and SiLU for Multi-branch WLU.}
\end{table}

As shown in Table \ref{tab: activation}, different activation functions lead to variations in model performance. Among the single activation functions, PReLU achieves the highest accuracy of 0.8218, followed closely by SiLU at 0.8190, then ELU and ReLU. In terms of AUC, SiLU leads with 0.7762, with the others trailing in the sequence provided. However, the weights assigned to the Multi-branch WLU components, including ELU, PReLU, and SiLU, significantly influence the overall performance. Our InterpreTabNet, with a weight allocation for ELU (0.6), PReLU (0.2), and SiLU (0.2) in the Multi-branch WLU, outperforms all other configurations. It achieves the highest Accuracy of 0.8252 and the highest AUC of 0.7814. This experiment highlights the importance of choosing the right activation functions and appropriately tuning the weights for the Multi-branch WLU to maximize the model's performance. The InterpreTabNet configuration demonstrates the effectiveness of our proposed approach in leveraging these components to achieve superior results.

\subsection{Comparative Experiment on Different Models}

Table \ref{tab: modelexperiment} summarizes the comparative results of various machine learning and deep learning models in terms of Accuracy and AUC.

\begin{table}[h]
    \begin{tabular}{ccc}
        \toprule
        Model & Accuracy & AUC  \\
        \midrule
        Logistic Regression & 0.7787 & 0.6414 \\
        SVM & 0.7788 & 0.5800 \\
        Random Forest & 0.8150 & 0.7629  \\
        MLP Neural Network & 0.7295 & 0.6412  \\
        LightGBM & 0.8185 & 0.7768  \\
        TabNet & 0.8205 & 0.7661  \\
        InterpreTabNet & \textbf{0.8252} & \textbf{0.7814} \\
        \bottomrule
    \end{tabular}
    \caption{Performance of different classifiers based on given evaluation metrics. The highest values are bold}
    \label{tab: modelexperiment}
\end{table}

Table \ref{tab: modelexperiment} provides an overview of the performance of various classifiers on the task at hand. While Logistic Regression achieves an Accuracy of 0.7787 and an AUC of 0.6414, SVM exhibits slightly better Accuracy (0.7788) at the cost of a lower AUC (0.5800). On the other end of the spectrum, MLP Neural Network lags behind with an Accuracy of 0.7295 and an AUC of 0.6412. Notably, LightGBM delivers strong performance, surpassing most models, with an Accuracy of 0.8185 and an AUC of 0.7768. TabNet demonstrates its competitive advantage with an Accuracy of 0.8205 and an AUC of 0.7661, indicating its effectiveness for the given task. However, our proposed InterpreTabNet stands out as the top-performing model, achieving the highest Accuracy of 0.8252 and the highest AUC of 0.7814, offering a balanced trade-off between Accuracy and AUC compared to the other models.

\subsection{Experiment for the Stability of Models}
\subsubsection{Validation for the Effectiveness of InterpreStability}

To validate the effectiveness of the proposed InterpreStability metric, we conduct experiments on datasets with varying data volumes and use Sklearn's permutation importance calculation method to obtain the permutation importance of InterpreTabNet on these datasets. We also compute the variance of permutation importance, along with InterpreStability values on the same datasets. Both metrics are employed to assess the stability of explaining feature importance, where higher InterpreStability and lower permutation importance variance indicate consistent model performance in explaining features and also means that our proposed evaluation metric has valid effectiveness for describing the stability of the model's interpretability.

Table \ref{tab: validation_datasets} summarizes the results obtained from different datasets. The data volumes of these datasets vary significantly, from 569 samples in the ``Breast Cancer Wisconsin" dataset to 30,000 samples in the ``Default of Credit Card Clients" dataset. The observed relationship between data volume and InterpreStability highlights the metric's ability to assess stability across datasets of varying sizes.

\begin{table}[h]
    \begin{tabular}{cccc}
        \toprule
        Dataset & Data Volume & InterpreStability & Permutation Importance Variance  \\
        \midrule
        Default of Credit Card Clients & 30,000 & \textbf{0.9772} & \boldmath{$2.027 \times 10^{-4}$} \\
        Breast Cancer Wisconsin & 569 & 0.2278 & $1.077 \times 10^{-3}$ \\
        Digits & 1,797 & 0.2111 & $1.503 \times 10^{-3}$ \\
        \bottomrule
    \end{tabular}
    \caption{InterpreStability and permutation importance variance values based on different datasets. The best values are highlighted in bold.}
    \label{tab: validation_datasets}
\end{table}

Table \ref{tab: validation_datasets} demonstrates the effectiveness of the InterpreStability metric. The significantly higher InterpreStability value for the ``Default of Credit Card Clients" dataset, combined with the notably lower permutation importance variance, underscores the stability of our model's feature explanation. Meanwhile, although the value of InterpreStability varies along with that of permutation importance variance, these values do not have a rigorous linear relationship with the dataset scale. These results offer empirical support for the effectiveness of our proposed metric in evaluating the stability of feature importance explanations.

\subsubsection{Experiment for the Stability of InterpreTabNet's Interpretability}
In order to test the stability of our model's interpretability, we employed both InterpreTabNet and TabNet models to perform feature extraction on several identical subsets of the training dataset. The specific steps are as follows:

\begin{enumerate}
    \item Set a random seed for data sampling and randomly extracted subsets of 2,333, 5,000, and 10,000 data points from the original 21,000 training samples, providing small, medium, and large-scale datasets, respectively.
    
    \item Another random seed is set for training the models. Nine different models are trained based on the three different scales of datasets mentioned above.
    
    \item Each model generates its set of feature importance scores, and we record these scores for each feature in the dataset.
    
    \item Pearson correlation coefficients are computed between each pair of feature importance scores generated by the nine models for each dataset. This results in a \(9 \times 9\) correlation matrix for each dataset.
    
    \item Based on these correlation matrices, we compute the InterpreStability score following the prescribed methodology.
\end{enumerate}

\begin{figure}[h]
  \centering
  \includegraphics[width=1\textwidth]{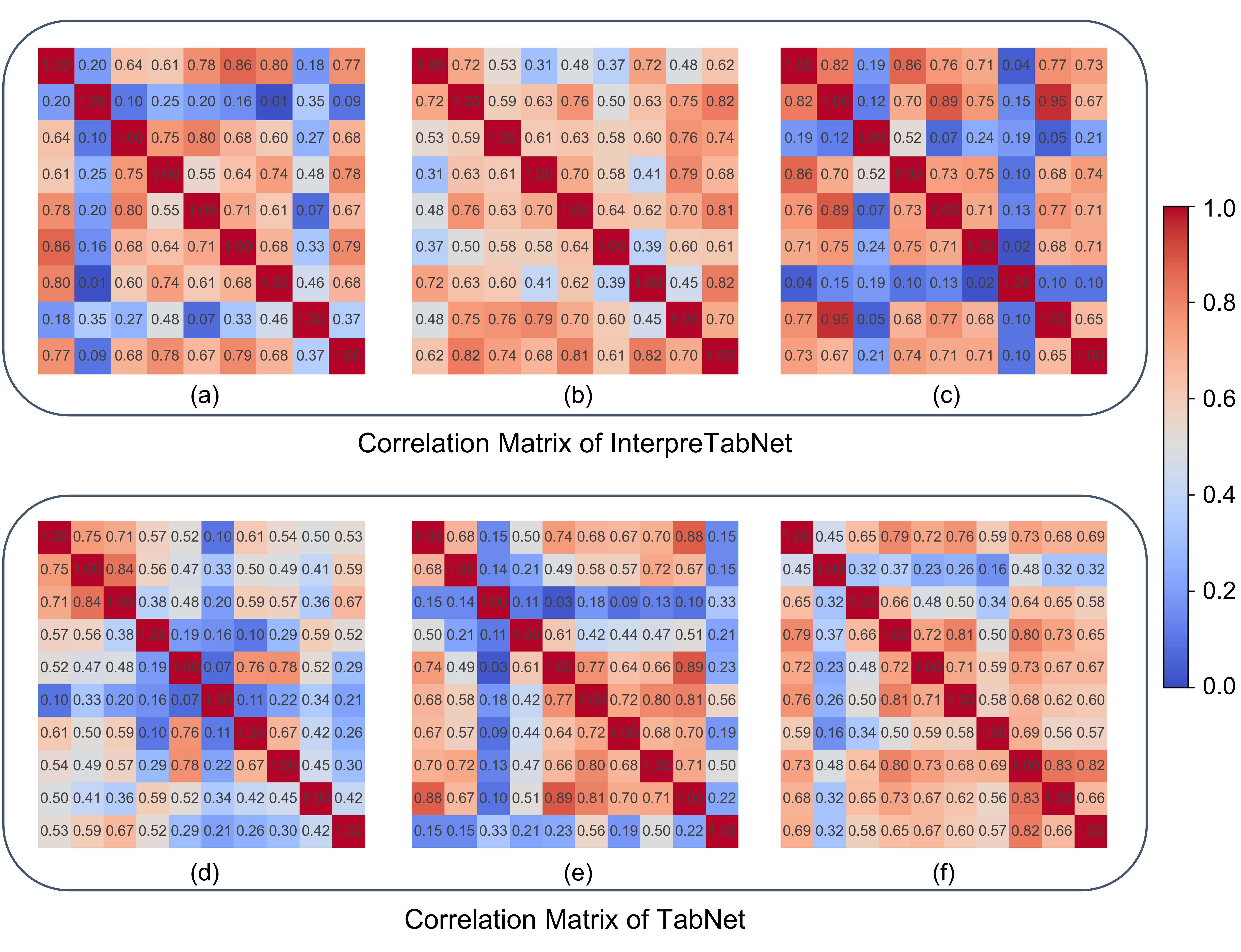}
  \caption{The visualization of correlation matrixes of two models based on three datasets. The upper part is InterpreTabNet and the lower part belongs to TabNet. (a) and (d) are based on 2,333 data points, (b) and (e) are based on 5,000 data points, (c) and (f) are based on 10,000 data points. The right color bar represents the strength of correlation, red color means high correlation, while blue color denotes low correlation.}
  \label{fig: interprestability}
\end{figure}

The correlation matrix of each model based on three different datasets is shown in Figure \ref{fig: interprestability}.

As illustrated in Figure \ref{fig: interprestability}, each heatmap forms a \(9 \times 9\) square, where each color block represents a set of feature importance generated by a model. The color gradient represents the degree of correlation between the feature importance sets generated by two different models. A warmer, more red hue indicates a higher correlation, signifying greater stability in feature selection by the model. In contrast, a cooler, more blue hue indicates less stability. It is noteworthy that the diagonal of each heatmap is the deepest red because the correlation of a feature importance set with itself is naturally 1. In comparing the heatmaps for InterpreTabNet and TabNet, the former shows a greater abundance of red blocks, highlighting its superior stability in feature selection. This improved performance is associated with our enhancements in the MLP-Attentive Transformer module. In the heatmaps for TabNet, there is a clear prevalence of blue blocks. For example, in sub-figure Figure \ref{fig: interprestability} (e), the third feature importance set shows a deep blue color, indicating a low correlation with other sets. From a horizontal perspective, the stability of both models increases as the dataset volume grows. This is related to the model's capability to select features in smaller datasets. When the dataset is small, the features acquired in each instance are sparsely distributed, making it difficult to measure the model's stability effectively.

The detailed experimental results are summarized in Table \ref{tab: stability_experiment}.

\begin{table}[h]
    \begin{tabular}{cccccccc}
        \toprule
        & & \multicolumn{5}{c}{Correlation coefficient distribution} &  \\ \cmidrule{3-7}
        Model & Data volume & [0, 0.3] & [0.3, 0.5] & [0.5, 0.7] & [0.7, 0.9] & [0.9, 1] & InterpreStability \\
        \midrule
                & 2,333 & 0.2778 & 0.1389 & 0.3056 & 0.2778 & 0 & 0.5167  \\
        InterpreTabNet & 5,000 & 0 & 0.2222 & 0.4167 & 0.3611 & 0 & \textbf{0.6278}  \\
               & 10,000 & 0.1944 & 0.1111 & 0.1389 & 0.5278 & 0.0278 & 0.6167  \\
        \midrule
               & 2,333 & 0.2889 & 0.2667 & 0.3333 & 0.1111 & 0 & 0.4533  \\
        TabNet & 5,000 & 0.3333 & 0.1556 & 0.2889 & 0.2222 & 0 & 0.4667  \\
               & 10,000 & 0.0667 & 0.1778 & 0.4889 & 0.2667 & 0 & \textbf{0.5911} \\
        \botrule
    \end{tabular}
    \caption{Summary of correlation coefficient distribution and InterpreStability for different models and dataset volumes. The highest InterpreStability value in each model is bold.}
    \label{tab: stability_experiment}
\end{table}

As shown in Table \ref{tab: stability_experiment}, the correlation coefficient distributions provide a multi-faceted perspective on model interpretability and stability across varying dataset volumes.

\begin{enumerate}
    \item A longitudinal analysis reveals a noteworthy trend in both models: as the dataset volume increases, the overall InterpreStability score also tends to rise, albeit non-linearly. For InterpreTabNet, the InterpreStability score starts at 0.5167 with 2,333 data points, peaks at 0.6278 for 5,000, and then drops slightly to 0.6167 at 10,000. In contrast, TabNet's score consistently rises: 0.4533 (2,333 points), 0.4667 (5,000 points), and 0.5911 (10,000 points). This could indicate a more gradual but steadier benefit from larger datasets.

    \item When we compare the two models laterally at each dataset volume, InterpreTabNet consistently outperforms TabNet in terms of InterpreStability. Especially at 5,000 and 10,000 data points, the InterpreStability score for InterpreTabNet is noticeably higher, emphasizing its more stable feature selection.

    \item The distributions of the correlation coefficients also offer insights. InterpreTabNet, for instance, has higher proportions of moderate to high correlation scores \([0.5, 0.7]\) and \([0.7, 0.9]\) as dataset volume increases, especially at 10,000 data points where the highest segment (\([0.7, 0.9]\)) dominates. TabNet, conversely, shows a much more spread distribution across lower and moderate correlation scores, with no particular dominance in higher segments.

    \item Overall, the results suggest that InterpreTabNet offers a more robust and stable feature selection across varying dataset volumes, likely due to its architectural improvements in the MLP-Attentive Transformer module. TabNet, however, appears to be more volatile, showing greater shifts in correlation distribution and InterpreStability, especially at the 10,000 data point level where its low correlation scores \([0, 0.3]\) drop significantly. Besides, the optimal dataset size for achieving the highest level of interpretability can vary between models.
\end{enumerate}

In addition to comparing InterpreTabNet with TabNet, we validate the stability of InterpreTabNet's interpretability using Random Forest and LightGBM. The experiment is performed on the ``Default of Credit Card Clients" dataset, with a substantial data volume of 30,000, which offers greater statistical power to detect differences in interpretability stability. Consequently, the observed variability in InterpreStability values among the models is more pronounced, allowing for a more nuanced understanding of model stability differences. We partition the dataset with different random seeds and utilize these partitions to obtain multiple sets of feature importance and permutation importance variance.

Table \ref{tab: validation_models} presents the results:

\begin{table}[h]
    \begin{tabular}{cccc}
        \toprule
        Model & InterpreStability & Permutation importance variance  \\
        \midrule
        Random Forest & \textbf{0.9928} & \boldmath{$1.913 \times 10^{-4}$} \\
        InterpreTabNet & 0.9772 & $2.027 \times 10^{-4}$  \\
        LightGBM & 0.9720 & $2.186 \times 10^{-4}$ \\
        \bottomrule
    \end{tabular}
    \caption{InterpreStability and permutation importance variance for different models. The best values are highlighted.}
    \label{tab: validation_models}
\end{table}

The experiment demonstrates that Random Forest exhibits the highest stability in explaining features, as indicated by the highest InterpreStability score (Random Forest $>$ InterpreTabNet $>$ LightGBM). Correspondingly, the permutation importance variance for the three models increases in the same order (Random Forest $<$ InterpreTabNet $<$ LightGBM), reaffirming the effectiveness of these evaluation metrics in describing interpretability stability. Furthermore, our InterpreStability metric exhibits a notable variability, making it easy to discern differences in model stability. Importantly, this metric is straightforward to implement and provides a clear indication of interpretability stability.

\section{Discussion} \label{sec6}
\subsection{Performance of InterpreTabNet based on Different Datasets}
In this study, we benchmarked a variety of machine learning and deep learning models, discovering that InterpreTabNet outperformed all other baseline models in both Accuracy and AUC metrics. This includes advanced tree-based models such as Random Forest and LightGBM. The superiority of InterpreTabNet can be attributed to three key improvements in its structure: 

\begin{enumerate}
    \item Firstly, by introducing an MLP structure into the Attentive Transformer module, InterpreTabNet has amplified its feature extraction and representation capabilities. In contrast, traditional models like Logistic Regression and SVM lack the depth to delve into the complex interactions among features, whereas the MLP in InterpreTabNet is adept at capturing these intricate nonlinear relationships.
    \item Secondly, the novel Multi-branch WLU provides a more potent expressiveness. This contrasts with traditional neural networks, such as the MLP Neural Network, which rely on common activation functions like ReLU or Sigmoid. Multi-branch WLU not only bolsters the model's expressivity but also enhances its computational stability and gradient propagation effectiveness.
    \item Lastly, with the Sparsemax activation function, InterpreTabNet maintains sparsity while augmenting its expressive power. Although other models like Random Forest and LightGBM can automatically select features, they do not directly account for the sparsity in activations.
\end{enumerate}

In terms of interpretability, 
\begin{enumerate}
    \item InterpreTabNet, when compared to TabNet, showcases improvements in both stability and robustness, with marginal discrepancies in InterpreStability values across different subsets of the same dataset. 
    \item While the InterpreStability of our model does vary significantly across datasets of different sizes (0.9772 for 30,000 data points and 0.2278 for 569 data points), this can be attributed to the vast difference in data volume, leading to pronounced shifts in the performance of deep learning models. However, when evaluated on smaller datasets, the InterpreStability values are more consistent (0.2278 for 569 data points and 0.2111 for 1,797 data points).
    \item When assessing the interpretability of different models on a large dataset, we observe that although Random Forest slightly leads in this metric (0.9928), both InterpreTabNet (0.9772) and LightGBM (0.9720) exhibit comparable performances. This can be caused by the fundamental differences between InterpreTabNet, a deep learning model, and the tree-based models, Random Forest and LightGBM. These models inherently differ in their internal structures and operational methodologies, which may influence their feature importance stability. Specifically, deep learning models might require more training iterations for convergence compared to their tree-based counterparts. This difference in the training process might lead to variations in the stability of feature importance. In feature selection, the ensemble nature of the Random Forest, averaging results across numerous decision trees, might lend to more stable feature importance. In contrast, LightGBM, with its gradient boosting mechanism and optimized feature selection process, offers efficiency. InterpreTabNet, leveraging attention mechanisms to select features, can easily capture subtle data nuances, potentially leading to fluctuations in feature importance when the dataset experiences perturbations. However, when considering both accuracy and interpretability, within an acceptable margin of error, our model remains the most balanced choice.
\end{enumerate}

\begin{figure}[h]
  \centering
  \includegraphics[width=1\textwidth]{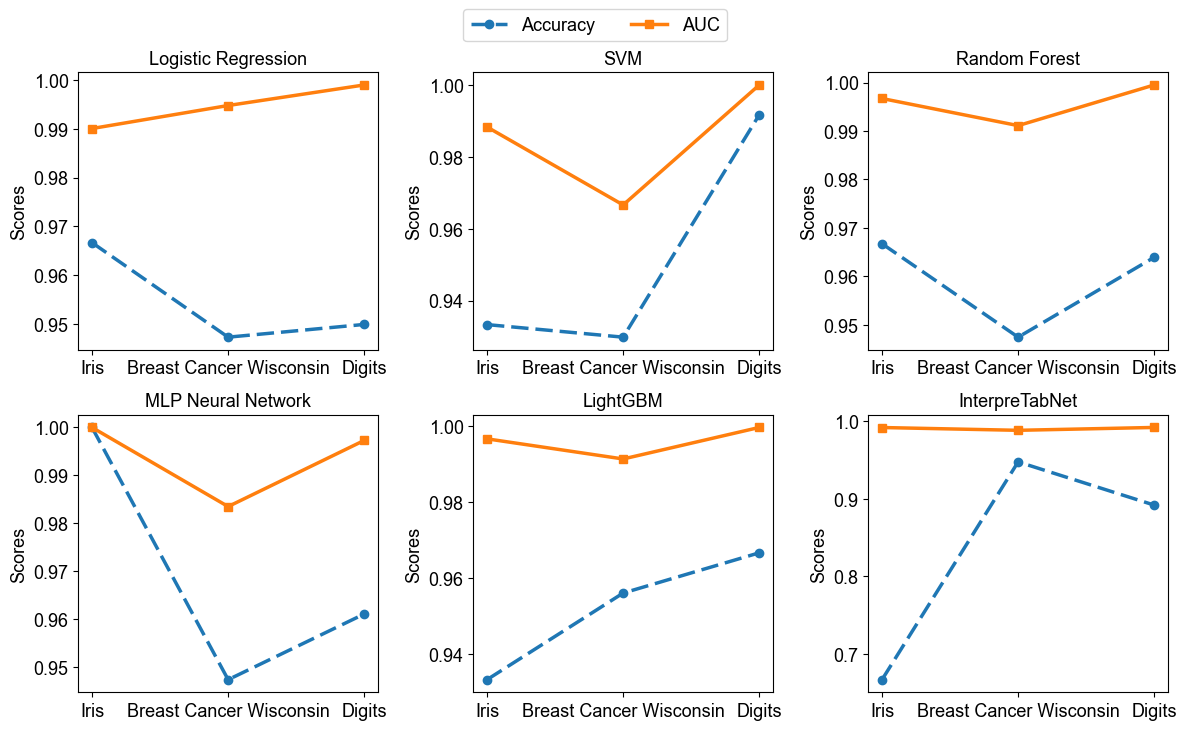}
  \caption{Comparative performance of models across three small datasets. The orange line with squared points denotes AUC and the blue dash line with circular points depicts the Accuracy.}
  \label{fig: discussion_performance}
\end{figure}

To further delve into the performance of the deep learning-based InterpreTabNet, we evaluated the performance of InterpreTabNet alongside other models on three small datasets with varying data volumes. The results are summarized in Table \ref{tab: performance_small} and visualized in Figure \ref{fig: discussion_performance}.

\begin{table}[h]
    \centering
    \begin{tabular}{ccccc}
    \toprule
    Dataset & Data Volume & Model & Accuracy & AUC \\
    \midrule
    \multirow{6}{*}{Iris} & \multirow{6}{*}{150} & Logistic Regression & 0.9667 & \textcolor{blue}{0.9900} \\
    &  & SVM & 0.9333 & \textcolor{blue}{0.9883}  \\
    &  & Random Forest & 0.9667 & 0.9967 \\
    &  & MLP Neural Network & \textcolor{red}{1.0000} & \textcolor{red}{1.0000} \\
    &  & LightGBM & 0.9333 & 0.9967 \\
    &  & InterpreTabNet & \textbf{\textit{0.6667}} & \textbf{\textit{0.9917}} \\
    \midrule
    \multirow{6}{*}{Breast Cancer Wisconsin} & \multirow{6}{*}{569} & Logistic Regression & 0.9474 & 0.9947 \\
    &  & SVM & \textcolor{blue}{0.9298} & \textcolor{blue}{0.9666}  \\
    &  & Random Forest & 0.9474 & 0.9911 \\
    &  & MLP Neural Network & 0.9474 & \textcolor{blue}{0.9835} \\
    &  & LightGBM & 0.9561 & 0.9914 \\
    &  & InterpreTabNet & \textbf{\textit{0.9474}} & \textbf{\textit{0.9881}} \\
    \midrule
    \multirow{6}{*}{Digits} & \multirow{6}{*}{1,797} & Logistic Regression & 0.9500 & 0.9989 \\
    &  & SVM & 0.9917 & \textcolor{red}{1.0000} \\
    &  & Random Forest & 0.9639 & 0.9995 \\
    &  & MLP Neural Network & 0.9611 & 0.9973 \\
    &  & LightGBM & 0.9667 & 0.9997 \\
    &  & InterpreTabNet & \textbf{\textit{0.8917}} & \textbf{\textit{0.9918}} \\
    \bottomrule
    \end{tabular}
    \caption{InterpreTabNet and other models' performance on three small datasets. Values of InterpreTabNet are marked in bold and italic. Values that are inferior to those for InterpreTabNet are marked blue. Outliers are marked red.}
    \label{tab: performance_small}
\end{table}

We conducted experiments on small datasets to delve deeper into the models' generalization and robustness across different scenarios. As a deep learning model, InterpreTabNet generally necessitates a substantial amount of data to achieve its full potential. However, the visual representation in Figure \ref{fig: discussion_performance} showcases InterpreTabNet's robustness even on smaller datasets. For instance, on the Iris dataset, despite its limited data volume, InterpreTabNet demonstrated commendable performance, surpassing some traditional machine learning models. 

A striking observation from Figure \ref{fig: discussion_performance} is the consistency of the AUC values for InterpreTabNet across datasets of varying sizes. This suggests that the model is effective in ranking predictions and distinguishing between the classes. On the other hand, the Accuracy of InterpreTabNet shows noticeable fluctuations as the dataset size changes. This can be attributed to the fact that AUC is a metric that evaluates the model's ability to discriminate between positive and negative classes without being tied to a specific threshold, while Accuracy is sensitive to the threshold chosen for classification. In situations with imbalanced datasets or limited data like these three datasets, slight changes or misclassifications can lead to significant shifts in Accuracy. 

It is also worth noting the overfitting experienced by the MLP Neural Network on the Iris dataset, likely attributed to the dataset's minimal size. In stark contrast, InterpreTabNet, after meticulous hyperparameter tuning, displayed consistent performance, a testament to its in-built sparse and regularization modules like Entmax and GBN. Though InterpreTabNet might not have clinched the top spots across these small datasets, its consistently close performance to the top-tier baselines underscores its versatility and adaptability across diverse scenarios.

\subsection{Feature Engineering in terms of Selected Features}
In our experiments, we employed the InterpreTabNet model to help us identify critical features in the test dataset. InterpreTabNet's decision steps were recorded, and the frequency of each feature being selected was tallied. Based on these feature selection frequencies, we reordered the attributes of the original dataset. High-importance features, as determined by their selection frequency, were moved to the front, allowing us to sort the attributes from left to right based on their importance. Subsequently, the dataset was re-partitioned following the same splitting rules, and the models were retested.

Surprisingly, while other baseline models like Logistic Regression, Random Forest, LightGBM, and MLP Neural Network showed improved performance on the new dataset, InterpreTabNet's performance deteriorated. This phenomenon could indicate that while the reordered features may have simplified the learning process for traditional models, InterpreTabNet possibly became over-optimized or too fitted to the specific feature order, causing it to perform worse. It suggests that although feature importance can improve model generalizability, the same might not be true for models that have inherent feature selection mechanisms, such as InterpreTabNet. Thus, care should be exercised when employing feature engineering strategies across different types of models. Future work may involve exploring deeper into understanding the nuances of feature selection across various algorithms, particularly in context-dependent scenarios.

\subsection{Limitations and Expectations}
While InterpreTabNet has shown significant promise, it is essential to acknowledge its limitations and outline potential areas for improvement.

Firstly, training InterpreTabNet on large-scale datasets can be time-consuming due to its neural architecture. Future work should focus on optimizing computational efficiency and developing lightweight versions of the model. This will make InterpreTabNet more accessible for real-world applications, particularly when computational resources are limited.

Additionally, there is room for improving the model's performance further. Although InterpreTabNet demonstrates satisfactory results in terms of accuracy, interpretability, and generalization, there are scenarios where its accuracy can be enhanced. Investigating advanced training techniques, exploring novel model architectures, and leveraging larger datasets can contribute to achieving higher accuracy levels. Furthermore, there is potential for further breakthroughs in the stability of the model's interpretability. Research efforts should also aim to discover more stable model variants or develop specialized stability-enhancing techniques. Meanwhile, the consistency of models' interpretability should be further explored.

In conclusion, InterpreTabNet represents a significant step towards interpretable and high-performing deep learning models. Recognizing its limitations and addressing them is significant for its continued development and broader adoption in various domains.

\section{Conclusion} \label{sec7}
In this study, we have successfully designed a deep learning model that combines high classification accuracy with stable interpretability for tabular data analysis. Our contributions encompass several key aspects:

\begin{enumerate}
    \item We enhanced the TabNet model by introducing an MLP structure within the Attentive Transformer module, significantly improving its feature extraction capabilities. This enhancement formed the foundation for our proposed model, InterpreTabNet, which achieved outstanding performance.
    \item We introduced a novel activation function called Multi-branch WLU, which enhanced the model's expressiveness and improved gradient propagation and computational stability. Multi-branch WLU empowered our model to capture complex relationships, leading to improved generalization.
    \item Replacing the Sparsemax activation function with the Entmax sparse activation function allowed us to strike a balance between sparsity and expressive power. This change facilitated the extraction of more relevant features, enhanced gradient propagation, and avoided overfitting, ultimately boosting model performance.
    \item To evaluate interpretability stability, we introduced the InterpreStability metric. This metric provides a comprehensive measure of how stable the model explains feature importance across various scenarios.
\end{enumerate}

Our comprehensive experiments across diverse datasets have consistently illustrated that InterpreTabNet not only achieves high accuracy but also excels in offering robust interpretability. This positions it as a potent tool for in-depth tabular data analysis. Moreover, our research endeavors offer actionable insights and guidelines for the construction of explainable AI models, specifically deep learning-based models. We also introduce innovative methodologies to gauge the interpretability of these models.

As we look forward, we are optimistic about introducing more enhancements to InterpreTabNet. Additionally, we aim to develop further metrics to enrich the domain of interpretable learning. Our primary aspiration is to catalyze a movement towards more transparent and efficient data mining and analytics across various application spectrums.

\section*{Acknowledgements}
We extend our appreciation to the University of California Irvine Machine Learning Repository and Scikit-learn for their data support. We utilized ChatGPT for refining the manuscript and debugging our code. We are grateful for the rapid advancements in artificial intelligence, which have significantly enhanced our work efficiency. We are also grateful to the reviewers and editors for their insightful feedback and recommendations. A heartfelt thank you to our corresponding author for her guidance and mentorship.

\section*{Authors' contributions}
Conceptualization: SW; Methodology: SW and XL; Data collection and analysis: SW and XL; Experimental design: SW and MW; Formal analysis and investigation: SW; Visualization: SW; Writing - original draft preparation: SW; Writing - review and editing: SW, XL and MW; Supervision: MW and SW. All authors read and approved the final manuscript.

\section*{Declarations}
\begin{itemize}
\item Funding:

The authors declare that no funds, grants, or other support were received during the preparation of this manuscript.

\item Conflict of interest:

The authors have no relevant financial or non-financial interests to disclose.

\item Ethics approval:

Not applicable.

\item Consent to participate:

Not applicable.

\item Consent for publication:

Not applicable.

\item Availability of data and materials:

All the data used in this paper are collected from the UCI Machine Learning Repository and Scikit-learn, which are public data resources.
\end{itemize}

\bigskip

\bibliography{sn-bibliography}

\end{document}